\documentclass[10pt,twocolumn]{article}
\usepackage[letterpaper,margin=0.75in]{geometry}
\usepackage{times}
\usepackage{xcolor}
\usepackage{amsmath,amssymb,amsfonts}
\usepackage{algorithmic,multirow,graphicx,textcomp,tikz,booktabs,subcaption,overpic,bm}
\usepackage{cite}
\usepackage{xspace}
\usepackage[colorlinks=true,allcolors=black]{hyperref}
\usepackage{authblk}

\newcommand{\figref}[1]{Fig.~\ref{#1}}
\newcommand{\tabref}[1]{Table~\ref{#1}}

\newcommand{\myparagraph}[1]{\emph{#1:}}
\newcommand{\etal}{et~al.\xspace}

\title{Average Calibration Losses for Reliable Uncertainty in Medical Image Segmentation}
\author[1]{Theodore Barfoot}
\author[1]{Luis C. Garcia-Peraza-Herrera}
\author[2]{Samet Akcay}
\author[3]{Ben Glocker}
\author[1]{Tom Vercauteren}
\affil[1]{King's College London (theodore.d.barfoot@kcl.ac.uk)}
\affil[2]{Intel Corporation}
\affil[3]{Imperial College London}
\date{}

\begin{document}
\twocolumn[
	\maketitle
	\begin{abstract}
		Deep neural networks for medical image segmentation are often overconfident, compromising both reliability and clinical utility.
		In this work, we propose differentiable formulations of marginal L1 Average Calibration Error (mL1-ACE) as an auxiliary loss that can be computed on a per-image basis.
		We compare both hard- and soft-binning approaches to directly improve pixel-wise calibration.
		Our experiments on four datasets (ACDC, AMOS, KiTS, BraTS) demonstrate that incorporating mL1-ACE significantly reduces calibration errors, particularly Average Calibration Error (ACE) and Maximum Calibration Error (MCE), while largely maintaining high Dice Similarity Coefficients (DSCs).
		We find that the soft-binned variant yields the greatest improvements in calibration, over the DSC plus cross-entropy loss baseline, but often compromises segmentation performance, with hard-binned mL1-ACE maintaining segmentation performance, albeit with weaker calibration improvement.
		To gain further insight into calibration performance and its variability across an imaging dataset, we introduce dataset reliability histograms, an aggregation of per-image reliability diagrams.
		The resulting analysis highlights improved alignment between predicted confidences and true accuracies.
		Overall, our approach provides practitioners with explicit control over the calibration-accuracy trade-off, enabling more reliable integration of deep learning methods into clinical workflows.
		We share our code here:
		\url{https://github.com/cai4cai/Average-Calibration-Losses}
	\end{abstract}
	\vspace{1em}
]


\section{Introduction}
\label{sec:introduction}

Deep neural networks (DNNs) often exhibit \emph{miscalibration}, meaning their predicted probabilities (confidence) do not reflect true outcome likelihoods (accuracy).
In practice, DNNs tend to be overconfident, especially when trained on limited data, as is common in medical imaging tasks~\cite{Guo2017, Mehrtash2020confidence}.
In the safety-critical domains of medical diagnosis and segmentation, this miscalibration poses serious risks.
A model might assign high confidence to an incorrect decision, misleading clinicians.
Conversely, well-calibrated models improve the \emph{trustworthiness} of AI predictions by aligning predictive confidence with accuracy~\cite{Guo2017, Dawood2023uncertainty}.
Better-calibrated models can support improved clinical decisions and patient outcomes.
For example, in radiotherapy planning, accurate confidence estimates help doctors identify uncertain regions in need of review~\cite{Bernstein_2021}.
Overall, calibration of DNNs is emerging as a key requirement for reliable deployment of AI in healthcare.

Historically, calibration has primarily been addressed using post-hoc methods.
Techniques like Platt scaling and temperature scaling (TS) apply simple corrections to a model's predictions after training~\cite{Guo2017}.
These approaches have limitations: they fail to leverage the model’s full capacity to learn calibration directly, and they usually employ global adjustments, unable to correct instance- or class-specific calibration errors.

To address this, we propose integrating calibration directly into the model's training (train-time calibration), enabling the network to intrinsically produce well-calibrated predictions without post-hoc adjustments.
Our method specifically employs the Average Calibration Error (ACE) as both a calibration metric and a differentiable auxiliary loss function.
Unlike the widely used Expected Calibration Error (ECE), which weights errors by bin population and is dominated by highly confident predictions~\cite{Bernstein_2021}, ACE treats all confidence ranges equally~\cite{Naeini2015}.
This uniform treatment is particularly important for medical segmentation tasks, where uncertain predictions often occur at clinically critical regions like tissue boundaries.

Calibration metrics traditionally rely on dataset-level statistics.
For classification tasks, this can lead to inefficient estimation from small mini-batches~\cite{Kumar2018MMCE, liang2020improved}.
However, segmentation outputs voxel-level predictions per image, inherently providing ample data for stable calibration estimates even from single images.
Exploiting this fact, we introduce a straightforward, differentiable ACE-based loss using standard probability binning operations, circumventing the need for soft relaxations or proxy losses~\cite{krishnan2020improving, Karandikar2021SoftCalib}.

To better understand calibration performance across image segmentation datasets, we introduce \emph{dataset reliability histograms}, an aggregation of per-image reliability diagrams. These histograms provide a comprehensive tool to visualise the calibration of a predictor over an entire dataset.

This work significantly extends our previous conference publication~\cite{Barfoot2024ACE}, introducing a soft-binned variant (sL1-ACE) with smoother gradients and substantially expanding our evaluation to four diverse medical imaging datasets.
Additionally, we conduct sensitivity studies on bin count and loss weighting, include the use of micro-averaged metrics and update our network architecture.
Throughout this work, we employ the popular Dice Similarity Coefficient (DSC) plus Cross-Entropy (CE) loss as our primary segmentation baseline~\cite{lossodyssey2021}.
Cross-entropy is a strictly proper scoring rule \cite{Gneiting2007Proper} and has been shown to yield better calibrated probabilities than DSC loss in medical image segmentation~\cite{Mehrtash2020confidence}.
Consequently, adding a CE term to DSC (DSC+CE) is expected to improve calibration relative to pure DSC training.

Our results show that incorporating mL1-ACE (both hard- and soft-binned) significantly reduces calibration errors compared to the baseline DSC+CE loss.
We evaluate calibration errors using both micro- and macro-averaging, and segmentation performance via DSC. While soft-binned mL1-ACE (sL1-ACE) achieves greater calibration improvements, it can slightly compromise segmentation accuracy. In contrast, hard-binned mL1-ACE (hL1-ACE) notably improves calibration while largely maintaining segmentation performance.

In summary, we propose \emph{mL1-ACE}, a differentiable calibration loss for semantic segmentation derived from marginal reliability diagrams.
Unlike previous classification-focused approaches, we leverage dense voxel-level predictions to efficiently compute calibration metrics per image.
Our method, evaluated on four public datasets, demonstrates the effectiveness of both binning strategies and introduces dataset reliability histograms for visualising large-scale calibration trends.

\section{Background and Related Work}%
\label{sec:background}%

\subsection{Uncertainty Estimation in Medical Segmentation}
Uncertainty in image segmentation predictions is commonly divided into \emph{aleatoric} (data-related) and \emph{epistemic} (model-related) uncertainty~\cite{Kendall2017}.
Several methods have been proposed to model these uncertainties.
Epistemic uncertainty is often estimated using \emph{Monte Carlo dropout}~\cite{Gal2016} or \emph{deep ensembles}~\cite{Lakshminarayanan2017}, while aleatoric uncertainty can be modelled by predicting per-voxel variances~\cite{Kendall2017} or through \emph{test-time augmentation}~\cite{Ayhan2018}.
More expressive probabilistic segmentation networks, such as the \emph{Probabilistic U-Net}~\cite{Kohl2018} and \emph{PHiSeg}~\cite{Baumgartner2019}, model a distribution over possible segmentations.
\emph{Stochastic Segmentation Networks (SSNs)}~\cite{Monteiro2020} extend this by modelling spatially correlated aleatoric uncertainty, enabling the generation of coherent, plausible segmentation variants.
The QU-BraTS challenge~\cite{Mehta2022}, part of the BraTS 2020 benchmark, provided a platform for evaluating these methods in brain tumour segmentation. Top-performing entries combined dropout, test-time augmentation, and ensembles to estimate voxel-level uncertainty and demonstrated that capturing both epistemic and aleatoric components improves calibration and reliability in medical segmentation tasks.
Nonetheless, in the majority of previous medical image segmentation works, little emphasis has been put on principled uncertainty estimation and calibration, and the need to consolidate uncertainty estimation towards established metrics.

It is important to note that methods modelling distributions over segmentations address a different aspect of uncertainty than confidence calibration. While distributional approaches capture epistemic and aleatoric uncertainty through multiple plausible outputs, confidence calibration focuses on ensuring that the predicted class probabilities from a single network output accurately reflect the likelihood of correctness. Both perspectives are valuable and complementary, however, our work addresses the latter, aiming to improve the reliability of softmax-derived confidence estimates.

\subsection{Reliability Diagrams and Calibration Metrics}

\begin{figure}[tb!]
    \centering
    \includegraphics[width=0.45\textwidth]{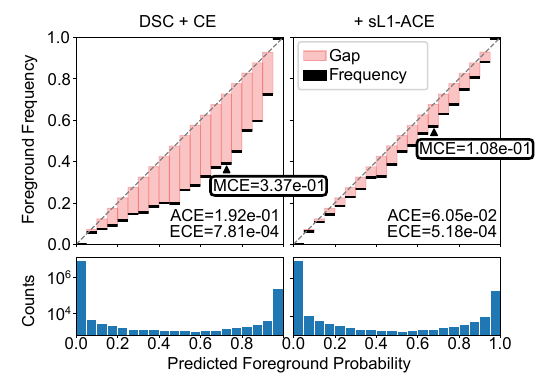}
    \caption{
        Reliability diagrams for BraTS 2021 (case 00095, whole tumour class), comparing baseline DSC plus CE loss (left) and baseline plus our auxiliary soft-binned mL1-ACE (sL1-ACE) loss (right).
        Each diagram shows empirical foreground frequency (accuracy) versus predicted foreground probability (confidence), voxel counts per bin, and associated calibration errors (ACE, ECE, MCE).
        The additional sL1-ACE loss substantially reduces the baseline model's overconfidence.
    }
    \label{fig:reliability_diagram_example}
\end{figure}

In the context of this work, we define calibration as \emph{confidence calibration}: the alignment between the model's predicted probability (confidence) and the true empirical accuracy. A model is considered well-calibrated if, for any given confidence level $p$, the proportion of correct predictions is approximately $p$.

Throughout this work, we use the terms \emph{confidence} and \emph{predicted probability} interchangeably, as both refer to the softmax output representing the model's belief in a particular class. Similarly, \emph{accuracy} and \emph{observed frequency} are used interchangeably to denote the empirical correctness rate. This deliberate equivalence reflects the reformulation of the traditional confidence-accuracy framework into a probability-frequency perspective suited to dense prediction tasks.

Reliability diagrams are a standard tool for visualising model calibration.
An example of such is shown in \figref{fig:reliability_diagram_example}.
They plot the actual accuracy achieved against the model predicted probability (confidence) for samples at that confidence level~\cite{DeGroot1983}.
To construct a reliability diagram, predicted class probabilities are discretised into bins.
For each bin, one computes the \emph{average confidence} of predictions in the bin (which will be close to the bin centre value) and the \emph{actual proportion} of those predictions that were correct.
The difference between the confidence and accuracy (DCA) in each bin is the calibration error for that bin.
A perfectly calibrated model would have zero difference for every bin.
In the context of classification tasks, a reliability diagram is typically constructed for a dataset.
However, in semantic segmentation, the output space is much larger, with each pixel/voxel representing a separate prediction.
This permits the construction of a reliability diagram for each image, meaning that derived calibration error metrics can be calculated on a per-image basis.
To the best of our knowledge, this observation has not been exploited in previous work.

Several scalar metrics distil calibration performance into a single number.
The most widely used is the Expected Calibration Error (ECE), which is a weighted average of the per-bin DCA, where the weights are the number of samples in each bin~\cite{Naeini2015}.
By contrast, the \emph{Average Calibration Error (ACE)} weights each bin equally by simply averaging the per-bin gaps~\cite{neumann2018a}.
The \emph{Maximum Calibration Error (MCE)} is another metric that reports the worst-case bin error~\cite{Naeini2015}.
Additional metrics, such as the Brier Score~\cite{Brier1950} and Overconfidence Error~(OE)~\cite{Thulasidasan2019OE}, have also been employed.
Most calibration metrics assess marginal or top-label calibration by evaluating each class independently.
In contrast, \emph{canonical} calibration requires that the predicted probability vector as a whole is calibrated with respect to the true class distribution.
This is more stringent and challenging to evaluate, especially in high-dimensional settings like segmentation.
Popordanoska~\etal~\cite{popordanoska2022canonical} addressed canonical calibration in classification by proposing a consistent and differentiable $L_p$ calibration error estimator based on Dirichlet kernel density estimation, enabling scalable, distribution-level calibration assessment.
While canonical calibration may offer insight beyond marginal calibration, it is beyond the scope of this work.

\subsection{Post-hoc Calibration Techniques}
One approach to mitigating DNN miscalibration is to address it after training in a post-hoc fashion, without modifying network weights.

A common approach is Temperature Scaling (TS)~\cite{Guo2017, platt1999probabilistic}, which multiplies logits by a scalar temperature to soften (temperature $>$ 1) or sharpen (temperature $<$ 1) predicted probabilities, and can be applied globally or per-class in segmentation tasks. Other methods include Platt scaling, isotonic regression, beta calibration, and Dirichlet calibration, with recent spatially aware extensions leveraging local neighbourhood information to further improve calibration~\cite{Rousseau2025,ding2020local}.

These methods provide a means to improve calibration, without affecting the accuracy of the predictor, but very few parameters are leveraged for this calibration, limiting the ability to provide context-aware and instance-specific calibration.

\subsection{Augmentation for Train-time Calibration}
An alternative to post-hoc calibration is \emph{train-time calibration} techniques, which leverage the high parameter count of DNNs to achieve calibration.
This can be performed indirectly; techniques such as \emph{label smoothing}, which softens hard one-hot labels, have been shown to reduce overconfidence and thereby improve calibration \cite{Szegedy2016}.
Additionally, data augmentation strategies, such as mixup~\cite{zhang2018mixup, Thulasidasan2019OE}, also contribute to model calibration by forcing the network to produce softer, more uncertain outputs on interpolated samples.

While these methods offer an implicit trade-off by reducing overconfidence but potentially degrading discriminative performance, they do not explicitly optimise a calibration metric and lack direct feedback on calibration quality during training.

\subsection{Auxiliary Calibration Losses in Classification}
A direct approach to improve model calibration is to modify the training objective by incorporating an auxiliary loss that explicitly targets calibration.
In image classification, several calibration-aware auxiliary losses have been proposed.
For instance, Liang~\etal~\cite{liang2020improved} introduced a loss term based on the discrepancy between the predicted probabilities and the empirical accuracy (derived from the expected calibration error, ECE), penalising the model when reductions in the cross-entropy loss do not yield corresponding improvements in accuracy.
According to Hebbalaguppe~\etal~\cite{hebbalaguppe2022stitch}, this DCA-based loss is purportedly limited by differentiability issues arising from discontinuities of the accuracy term due to the reliance on hard binning.
To address this potential issue, Hebbalaguppe~\etal~\cite{hebbalaguppe2022stitch} proposed the Multi-class DCA (MDCA) loss, which computes the calibration error by averaging over each class within a mini-batch rather than within fixed confidence bins.
Moreover, Kumar~\etal~\cite{Kumar2018MMCE} introduced the Maximum Mean Calibration Error (MMCE) loss, a differentiable formulation that can be optimised jointly with the primary task.
Other works have explored differentiable accuracy versus uncertainty calibration (AvUC) losses~\cite{krishnan2020improving} and soft calibration objectives (S-AvUC) that employ continuous binning to produce differentiable estimates~\cite{Karandikar2021SoftCalib}.
Recently, Bohdal~\etal~\cite{Bohdal2023MetaCalib} developed a differentiable surrogate for expected calibration error (DECE) using soft binning and smooth approximations, although its reliance on meta-learning increases complexity and may limit wider applicability.
As mentioned above, auxiliary losses have also been extended to canonical calibration~\cite{popordanoska2022canonical}.
Recent uncertainty-aware training strategies in medical image classification demonstrate that incorporating calibration-related loss terms can reduce miscalibration by a substantial margin while maintaining (or even improving) accuracy~\cite{Dawood2023uncertainty}.
Dawood~\etal showed that a \emph{Confidence Weight} loss penalty on incorrect high-confidence predictions reduced the expected calibration error by 17--22\% in cardiac MRI classification tasks, with slight accuracy gains~\cite{Dawood2023uncertainty}.

The approach of auxiliary calibration losses has significant limitations in the context of image classification due to the small output space, as each input image only has one associated prediction and therefore a single confidence value and a binary accuracy term.
This can be alleviated by using a larger mini-batch, but this is limited due to large input images, hence why proxy calibration objectives or efficient sampling methods have to be employed.

\subsection{Auxiliary Calibration Losses in Segmentation}
In contrast to classification problems, segmentation tasks produce dense, per-pixel/voxel predictions, yielding a much larger number of confidence estimates per image.
This can make calibration during mini-batch optimisation more effective, as the larger number of samples provides more stable and statistically meaningful estimates of calibration error for each forward pass.
The use of auxiliary calibration losses in semantic segmentation is less common than in classification, yet it is equally critical in medical imaging applications.
Standard segmentation losses, such as DSC loss, have been shown to produce overconfident predictions~\cite{Mehrtash2020confidence}, despite their effectiveness in handling class imbalance.
To address this, Yeung~\etal~\cite{Yeung2023DSC++} proposed the DSC++ loss, which modifies the DSC loss by adjusting false positive and false negative penalties according to class frequency.
Similarly, Neighbour-Aware Calibration (NACL) methods~\cite{Murugesan2023NACL,murugesan2024neighbor} incorporate spatial consistency by penalising overconfident predictions in local neighbourhoods.
Despite reported reduction in overconfidence with these approaches, there remains a need for a more controlled approach to ensure that model confidence estimates align more closely with observed accuracy while maintaining high segmentation performance.

\section{Method}
\label{sec:method}

\subsection{Reliability Diagrams and Dataset Reliability Histograms}
Reliability diagrams are constructed by discretising model predictions into probability bins and comparing the average predicted probability within each bin against the observed empirical frequency of that class. An illustrative example of a reliability diagram is provided in \figref{fig:reliability_diagram_example}.
We make the observation that in semantic segmentation, due to the highly dimensional prediction space, calibration errors can be computed on a per-image basis.
In our primary formulation, for each image and for each class \(c\), the predicted probability space is discretised into \(M\) bins using a hard-binning approach.
Then, for each bin \(m\), we compute the expected foreground probability \(e^c_m\) and the observed foreground frequency \(o^c_m\).
The difference between these two quantities represents the difference between confidence and accuracy (DCA) in a marginal, class-wise context.
More formally, we define:
\begin{equation}
	o^c_m = \mathbb{E}\!\left[\bm{Y}^c_i \,\middle|\, i \in B^c_m\right], \quad e^c_m = \mathbb{E}\!\left[f_{\boldsymbol{\theta}}(\bm{X})^c_i \,\middle|\, i \in B^c_m\right],
	\label{eq:o}
\end{equation}
where \(\bm{X}\) is the input image, or dataset (for micro-average metrics), \(\bm{Y}\) the associated ground truth label-map, \(B^c_m\) denotes the set of voxels \(i\) in the image whose predicted probability for class \(c\) falls into bin \(m\), and \(f_{\boldsymbol{\theta}}\) is our predictor (typically a ConvNet) with trainable weights \(\theta\).

This approach can trivially be applied across all voxels in the dataset (micro-averaging) to produce a reliability diagram for the dataset.
However, as we are able to construct reliability diagrams per-image,
we can leverage these to then construct a \emph{dataset reliability histogram} (rather than diagram).
This enables more fine-grained analysis of calibration performance across the dataset.
We start by collecting all \((e^c_m; o^c_m)\) pairs for all images.
For each bin $m$, by construction, the $e^c_m$ values across the images will be close to the centre of $B^c_m$.
The image-specific $o^c_m$ values, however, can vary significantly within each bin $m$.
To get insight into the corresponding distribution,
we compute the histogram of image-specific \(o^c_m\) values within each bin $m$.
The resulting dataset reliability histogram is visualised in 2D by plotting,
for each $m$ bin along the horizontal axis,
the 1D histograms of \(o^c_m\) values along the vertical axis.
As illustrated in \figref{fig:dataset_reliability_diagram}, this provides an intuitive visualisation of calibration variability across the entire dataset.

\subsection{Micro- and Macro-Averaged Calibration Metrics}
We use the sum of absolute differences between \(e^c_m\) and \(o^c_m\), which we refer to as L1 differences, as the basis of our calibration metrics:
\begingroup\allowdisplaybreaks
\begin{align}
	\textrm{mL1-ECE} & = \frac{1}{C} \sum_{c=1}^{C} \sum_{m=1}^{M} \frac{n^c_m}{N} \left| o^c_m - e^c_m \right|, \label{eq:ml1_ece} \\
	\textrm{mL1-ACE} & = \frac{1}{CM} \sum_{c=1}^{C} \sum_{m=1}^{M} \left| o^c_m - e^c_m \right|, \label{eq:ml1_ace}                \\
	\textrm{mL1-MCE} & = \frac{1}{C} \sum_{c=1}^{C} \max_{m} \left| o^c_m - e^c_m \right|, \label{eq:ml1_mce}
\end{align}
\endgroup
where \(n^c_m\) is the number of voxels in bin \(B^c_m\), \(C\) the number of classes, and \(N\) is the total number of voxels.

In our preliminary work~\cite{Barfoot2024ACE}, these calibration metrics were computed per image (macro-averaging) and then averaged over the dataset.
Here, we additionally perform micro-averaging, aggregating predictions
across all images.
Specifically, aggregated foreground probabilities are simple averages, observed frequencies are bin-count weighted averages, and bin counts are summed directly.
In our implementation, this aggregation is performed on-the-fly without storing the entire output tensor in memory, ensuring efficiency even for large datasets.

\subsection{Hard- and Soft-Binned ACE Loss Functions}
\begin{figure}[bh!]
	\centering
	\includegraphics[width=0.45\textwidth]{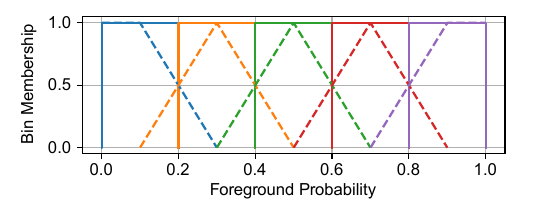}
	\caption{Visualisation of hard- (solid line) and soft- (dashed line) foreground probability binning assignment functions for 5 bins.}
	\label{fig:binning_comparison}
\end{figure}
For training, we employ the marginal L1 Average Calibration Error (mL1-ACE) loss defined in \eqref{eq:ml1_ace} in two variants: hard- and soft-binning, which we also refer to as hL1-ACE and sL1-ACE, respectively.
In this formulation, we define a \emph{membership function} \(\psi_m(x)\) that assigns a weight to a prediction \(x\) for bin \(m\), as shown in \figref{fig:binning_comparison}.
As is typical with kernel density estimation, the choice of kernel determines the nature of the binning.

\myparagraph{Hard-Binning (Square Kernel)}
For hard-binning, the membership function is a square wave:
\[
	\psi^{\text{hard}}_m(x) =
	\begin{cases}
		1, & \text{if } x \in [b_{m-1}, b_m), \\[1mm]
		0, & \text{otherwise},
	\end{cases}
\]
where \(b_0, b_1, \ldots, b_M\) are the bin boundaries defined by a uniform partition of the interval \([0,1]\); that is,
\[
	b_k = \frac{k}{M}, \quad k = 0, 1, \dots, M.
\]

\myparagraph{Soft Binning (Triangular Kernel)}
For soft-binning, we define a triangular kernel centred at the midpoint of each bin.
The membership function for bin \(m\) is:
\[
	\psi^{\text{soft}}_m(x) =
	\begin{cases}
		1,                         & \text{if } m = 1 \text{ and } x < \beta_1,    \\
		1 - M \cdot |x - \beta_m|, & \text{if } \beta_{m-1} \le x \le \beta_{m+1}, \\
		1,                         & \text{if } m = M \text{ and } x > \beta_M,    \\
		0,                         & \text{otherwise},
	\end{cases}
\]
where \(\beta_m = \frac{b_{m-1} + b_m}{2}\) is the centre of bin \(m\). Each triangular kernel peaks at 1 in its bin centre \(\beta_m\), decays linearly to 0 at the adjacent bin centres \(\beta_{m-1}\) and \(\beta_{m+1}\), and reaches 0.5 at the bin edges \(b_{m-1}\) and \(b_m\).

Then, for class \(c\), the expected foreground probability and observed foreground frequency are defined as
\begin{equation}
	e^c_m = \frac{\sum_{i \in B_m^c} \psi_m(x^c_i) \, x^c_i}{\sum_{i \in B_m^c} \psi_m(x^c_i)},
	\label{eq:soft_e_final}
\end{equation}
\begin{equation}
	o^c_m = \frac{\sum_{i \in B_m^c} \psi_m(x^c_i) \, Y^c_i}{\sum_{i \in B_m^c} \psi_m(x^c_i)},
	\label{eq:soft_o_final}
\end{equation}
where \(Y^c_i\) is the ground truth indicator for class \(c\) and \(x^c_i\) is the predicted probability for class \(c\) at voxel \(i\).
By adopting a kernel density estimation perspective, our method can easily be extended from using a square kernel (hard-binning), to a triangular kernel (soft-binning), or to any other suitable kernel (e.g. Spline, Gaussian, etc.) to compute voxel-level membership functions.


\section{Experiments}
\label{sec:experiments}

\subsection{Datasets}
We evaluate our auxiliary calibration losses on four public medical image segmentation datasets:
(1)~\textbf{ACDC 2017} -- the Automatic Cardiac Diagnosis Challenge Dataset \cite{ACDC2017}, a cardiac MRI dataset with left- and right-ventricle and myocardium segmentation classes;
(2)~\textbf{AMOS 2022} -- the large-scale Abdominal Multi-Organ Benchmark for Versatile Medical Image Segmentation \cite{AMOS2022}, an abdominal CT and MRI dataset with 15 organ classes, from which only the CT images were used;
(3)~\textbf{BraTS 2021} -- the Multimodal Brain Tumour Segmentation Benchmark dataset \cite{BraTS2021}, a multimodal neurological MRI dataset with three tumour classes; and
(4)~\textbf{KiTS 2023} -- the 2023 Kidney and Kidney Tumor Segmentation Challenge \cite{KiTS2023}, a CT with semantic classes for kidneys, renal tumours and renal cysts.
The ACDC, AMOS and KiTS datasets were chosen due to their reported suitability for benchmarking segmentation methods \cite{Isensee2024}, while BraTS 2021 was included as it is widely used (including our previous work \cite{Barfoot2024ACE}).

The ACDC dataset provides separate training and testing datasets, with $100$ cases in the training dataset and $50$ cases in the testing dataset. A training-validation split of $90$:$10$ was used, and the testing dataset was used as provided.
For AMOS, only the CT images were used.
The training dataset of 240 CT images was then split into training and validation with a ratio of $216$:$24$. The validation dataset was used as the testing dataset, as the testing dataset is not publicly available.
For the BraTS dataset, the publicly available training dataset of $1251$ cases was split into training, validation and testing with a ratio of $1000$:$51$:$200$ cases, using the same testing split as our previous work \cite{Barfoot2024ACE}.
For KiTS, the training dataset of $489$ cases was partitioned into training, validation and testing with a ratio of $352$:$39$:$98$.

Of note for evaluation, the AMOS testing set has 13\% of cases missing organs (e.g.\ gallbladder or prostate/uterus) due to field-of-view variation; BraTS has 4\% of cases lacking necrotic or enhancing tumour labels; and over half (56\%) of KiTS cases omit the cyst class.

\subsection{Training and Evaluation}
All training and evaluation was implemented via the MONAI framework \cite{MONAI}.
Our open-source code can be found at \url{https://github.com/cai4cai/Average-Calibration-Losses}.
A SegResNet model \cite{segresnet2019} was used, with 32 initial filters and 4 deep supervision layers, as employed by the KiTS 2023 competition winner \cite{Myronenko2024}.
We used the AdamW optimiser with a learning rate of \(1\times10^{-4}\), \(2\times10^{-4}\) for AMOS, and weight decay of \(1\times10^{-5}\).
A warmup cosine scheduler was used, with a linear warmup period of 2 epochs with a warmup multiplier of 0.50, followed by cosine annealing for the remaining epochs.
Training was run for 600 epochs for the ACDC and AMOS datasets and 1250 epochs for the BraTS and KiTS datasets.
Validation was performed every 10 epochs, with the best model saved based on the validation DSC score.
A fixed batch size of 2 was used, with the patch size maximised to fit within the GPU memory.

For the BraTS 2021 dataset, the same data augmentation preprocessing steps were applied as in \cite{Fidon2022}.
For the remaining datasets, a minimal preprocessing pipeline was used, consisting of random foreground cropping, resampling, random affine transformations and intensity normalisation.

%
%
For each dataset, we evaluated five configurations:
(1) a DSC+CE baseline with a 1:1 weight ratio between DSC and cross-entropy;
(2) a CE-only baseline trained with cross-entropy alone;
(3) DSC+CE with an additional hard-binned mL1-ACE term (hL1-ACE) using a 1:1:1 weighting for DSC, CE, and ACE;
(4) DSC+CE with a soft-binned mL1-ACE term (sL1-ACE), also with 1:1:1 weighting; and
(5) a temperature-scaled DSC+CE model obtained via post-hoc temperature scaling.

Temperature scaling was applied post-hoc to the DSC+CE baseline by optimising a single temperature parameter on the validation set for one epoch using cross-entropy and the Adam optimiser (learning rate 0.001).

For the auxiliary calibration losses, 20 bins were used to discretise the foreground probability space.
For any classes with no ground truth present, the mL1-ACE loss was set to zero during optimisation.

The saved model with the highest validation Dice Similarity Coefficient (DSC) was used for evaluation on the testing dataset. DSC, Average Calibration Error (ACE), Expected Calibration Error (ECE), and Maximum Calibration Error (MCE) were computed for each model, using both micro- and macro-averaging.
For all calibration metrics, a fixed number of 20 bins was used, with hard-binning of predicted foreground probabilities.
Additionally, reliability diagrams were generated for each testing case and a dataset reliability histogram for the whole dataset.

\subsection{Sensitivity Studies}
To assess the sensitivity of our auxiliary losses to hyperparameters, we conducted sensitivity studies on the ACDC dataset examining:
(1) the effect of bin count \(M\) on both hard and soft mL1-ACE variants, testing $M \in \{5, 10, 20, 50, 100\}$ bins, and
(2) the effect of loss weighting by varying the calibration loss weight $\lambda$ in the formulation DSC:CE:$\lambda$·ACE with fixed DSC:CE ratio of 1:1 and $\lambda \in \{0.1, 0.25, 0.5, 1.0, 2.0, 4.0, 10.0\}$ applied to the ACE component.
Note that $\lambda = 1.0$ corresponds to the 1:1:1 weighting (DSC:CE:ACE) used in all other experiments.

\section{Results}
\label{sec:results}
We report results, averaged across classes, excluding background, and for BraTS and KiTS we report the average of the hierarchical evaluation classes (HEC).
HECs reflect the labels used clinically and group together sub-regions into a larger class, such as whole tumour for BraTS, to create nested or composite labels.
We evaluate the performance of the baseline model, trained with DSC and cross-entropy (CE), against models trained with mL1-ACE auxiliary losses, with both hard and soft binning.
We compare the models across four datasets: ACDC, AMOS, BraTS, and KiTS.
We achieve a DSC score of 0.871 on ACDC, 0.883 on AMOS, 0.905 on BraTS, and 0.859 on KiTS with the baseline model.
This can be compared to state-of-the-art values of 0.927, 0.897, 0.915 and 0.887, respectively, as reported in the literature \cite{Isensee2024}.

\subsection{Tabular Results}
We find that both hard- and soft-binned mL1-ACE outperform the baseline in terms of macro- and micro-averaged ACE and MCE across all datasets.
In general, both variants are able to maintain DSC performance across datasets, though in some cases this may vary slightly.
%
%
To contextualise these results, we additionally compare our proposed mL1-ACE losses against a CE-only baseline. Compared to DSC+CE, training with CE only yields modest improvements in macro-ACE (around 3--6\% reduction across datasets) but consistently reduces DSC, illustrating the trade-off between segmentation accuracy and probabilistic calibration. Relative to CE only, Hard mL1-ACE further reduces macro-ACE by approximately 7--22\% on three of the four datasets while maintaining DSC close to the DSC+CE baseline, and Soft mL1-ACE achieves even larger macro-ACE reductions of roughly 16--44\% while preserving segmentation performance comparable to CE only (see Table~\ref{tab:dsc_ace_sig}).
%
%
We also compare against post-hoc temperature scaling applied to the DSC+CE baseline. Temperature scaling improves macro-ACE by roughly 8--25\% across datasets while maintaining DSC, confirming its effectiveness as a simple post-hoc calibration method. In comparison, Hard mL1-ACE attains calibration performance that is broadly comparable to temperature scaling with similar DSC, whereas Soft mL1-ACE further reduces macro-ACE by approximately 6--28\% relative to temperature scaling, providing the strongest overall calibration improvements among the evaluated methods (Table~\ref{tab:dsc_ace_sig}).

%
Notably, soft-binned mL1-ACE outperforms both hard-binned mL1-ACE and the baseline in terms of macro-averaged ACE (\tabref{tab:dsc_ace_sig}).
Specifically, soft mL1-ACE improves macro-averaged ACE by 46\% on ACDC, 19\% on AMOS, 30\% on BraTS and 36\% on KiTS compared to the baseline.
For micro-averaged ACE, soft mL1-ACE provides improvements of 70\% on ACDC, 52\% on AMOS, and 58\% on BraTS.
However, on KiTS, hard-binned mL1-ACE achieves the largest improvement, with a 58\% enhancement over the baseline, as shown in  \tabref{tab:dsc_ace_sig}.

\begin{table*}[t]
    \centering
    \caption{Table showing the Dice Similarity Coefficient (DSC) and Average Calibration Error (ACE) metrics with macro- and micro-averaging for DSC+CE baseline, CE only, temperature scaling, hard mL1-ACE, and soft mL1-ACE loss function models across four testing datasets.
        Temperature scaling is applied post-hoc to the DSC+CE baseline.
        Values show the average class metric, with BraTS and KiTS evaluated on the hierarchical class structure.
        Values in bold indicate the best performance for each dataset.
        Asterisks denote significance levels relative to the DSC+CE baseline: * p~\textless~0.05, ** p~\textless~0.01 (no symbol: non-significant).
        Standard deviation and statistical significance is not reported for micro-ACE as only one value is available for each dataset.}
    \label{tab:dsc_ace_sig}
    \begin{tabular}{llllc}
        \toprule
        \textbf{Loss Function} & \textbf{Dataset} & \textbf{DSC} $\uparrow$      & \textbf{Macro-ACE} $\downarrow$ & \textbf{Micro-ACE} $\downarrow$ \\
        \midrule
        \multirow{4}{*}{DSC + CE (Baseline)}
                               & ACDC 17          & \textbf{0.871 ± 0.037}       & 0.136 ± 0.029                   & 0.116                           \\
                               & AMOS 22          & \textbf{0.883 ± 0.044}       & 0.107 ± 0.025                   & 0.060                           \\
                               & BraTS 21         & \textbf{0.905 ± 0.107}       & 0.146 ± 0.070                   & 0.060                           \\
                               & KiTS 23          & 0.859 ± 0.144                & 0.171 ± 0.077                   & 0.110                           \\
        \midrule
        \multirow{4}{*}{CE Only}
                               & ACDC 17          & 0.868 ± 0.039$^{*}$          & 0.130 ± 0.028$^{**}$            & 0.109                           \\
                               & AMOS 22          & 0.872 ± 0.045$^{**}$         & 0.103 ± 0.024$^{**}$            & 0.053                           \\
                               & BraTS 21         & 0.895 ± 0.130$^{**}$         & 0.142 ± 0.063                   & 0.047                           \\
                               & KiTS 23          & 0.853 ± 0.139                & 0.160 ± 0.080                   & 0.139                           \\
        \midrule
        \multirow{4}{*}{DSC + CE + Temp. Scaling}
                               & ACDC 17          & \textbf{0.871 ± 0.037}       & 0.102 ± 0.033$^{**}$            & 0.073                           \\
                               & AMOS 22          & \textbf{0.883 ± 0.044}       & 0.093 ± 0.019$^{**}$            & 0.033                           \\
                               & BraTS 21         & \textbf{0.905 ± 0.107$^{*}$} & 0.135 ± 0.061$^{**}$            & 0.033                           \\
                               & KiTS 23          & 0.859 ± 0.144$^{*}$          & 0.145 ± 0.076$^{**}$            & 0.088                           \\
        \midrule
        \multirow{4}{*}{DSC + CE + Hard mL1-ACE (ours)}
                               & ACDC 17          & \textbf{0.870 ± 0.039}       & 0.101 ± 0.029$^{**}$            & 0.069                           \\
                               & AMOS 22          & 0.881 ± 0.047$^{*}$          & 0.103 ± 0.027$^{*}$             & 0.052                           \\
                               & BraTS 21         & \textbf{0.905 ± 0.112}       & 0.132 ± 0.064$^{**}$            & 0.050                           \\
                               & KiTS 23          & \textbf{0.863 ± 0.131}       & 0.131 ± 0.066$^{**}$            & \textbf{0.046}                  \\
        \midrule
        \multirow{4}{*}{DSC + CE + Soft mL1-ACE (ours)}
                               & ACDC 17          & 0.864 ± 0.044$^{**}$         & \textbf{0.073 ± 0.030$^{**}$}   & \textbf{0.035}                  \\
                               & AMOS 22          & 0.879 ± 0.043$^{**}$         & \textbf{0.087 ± 0.022$^{**}$}   & \textbf{0.029}                  \\
                               & BraTS 21         & 0.888 ± 0.127$^{**}$         & \textbf{0.102 ± 0.060$^{**}$}   & \textbf{0.025}                  \\
                               & KiTS 23          & 0.860 ± 0.130                & \textbf{0.109 ± 0.068$^{**}$}   & 0.085                           \\
        \bottomrule
    \end{tabular}
\end{table*}

For ACDC, statistically significant improvements (p~\textless~0.01) are observed in macro-ACE, macro-ECE, and macro-MCE when comparing the baseline to both variants; DSC is significantly lower only for soft mL1-ACE.
For the AMOS dataset, macro-ACE and MCE are improved significantly (p~\textless~0.05) by both mL1-ACE variants, with soft mL1-ACE showing the most significant improvements.
While the reduction in DSC, due to the introduction of the auxiliary loss functions, is small it is still statistically significant (p~\textless~0.01).
There is no statistically significant reduction in ECE for either mL1-ACE variant.
For BraTS, both calibration metrics macro-ACE and macro-MCE show significant differences (p~\textless~0.01) for both hard and soft L1-ACE compared to the baseline; DSC is significantly decreased only with soft mL1-ACE.
For KiTS, similar to the ACDC dataset, statistically significant improvements (p~\textless~0.01) in macro-ACE and macro-MCE are observed for both mL1-ACE variants. In this case, no significant differences in DSC or macro-ECE are observed.

\subsection{Radar Plots}
\begin{figure*}[htbp]
    \centering
    \includegraphics[width=\textwidth]{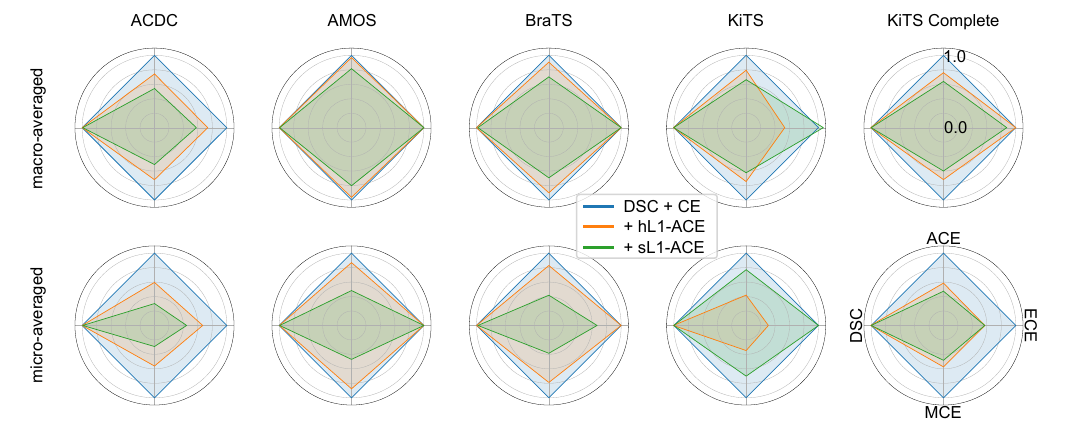}
    \caption{Normalised radar plots showing DSC, ACE, ECE and MCE across the four testing datasets and, in the rightmost column, for the KiTS testing dataset excluding cases with missing classes, keeping only complete cases. Models were trained with three different losses. Blue: baseline (DSC + CE), Orange: baseline plus hard-binned mL1-ACE (hL1-ACE), Green: baseline plus soft-binned L1-ACE (sL1-ACE).}
    \label{fig:normalized_spider_plots}
\end{figure*}

These observations are further supported by the normalised radar plots in  \figref{fig:normalized_spider_plots}, which show that both mL1-ACE variants are able to reduce calibration errors while largely maintaining DSC performance across all datasets.
The exception to this pattern is observed with the KiTS datasets, where soft mL1-ACE (sL1-ACE) exhibits an increased macro-ECE over baseline, and worse micro-averaged calibration performance compared to hard mL1-ACE (hL1-ACE).
However, this departure of the KiTS dataset from the overall trend is corrected when we evaluate the model on a version of the testing dataset that only contains cases with no missing classes.
The radar plot in this case is shown in the rightmost column of  \figref{fig:normalized_spider_plots}, where the soft-binned mL1-ACE outperforms the baseline and hard-binned variant for both macro- and micro-average calibration metrics.
The presence of missing classes, also occurs for the AMOS and BraTS datasets. When evaluating the models on the testing datasets with no missing classes, we observe a slight uplift in DSC performance of all models, however, it does not significantly change the relative calibration performance of the models, and ECE still shows limited improvements using mL1-ACE auxiliary losses.

\subsection{Dataset Reliability Histograms}
\begin{figure*}[htbp]
    \centering
    \includegraphics[width=\textwidth]{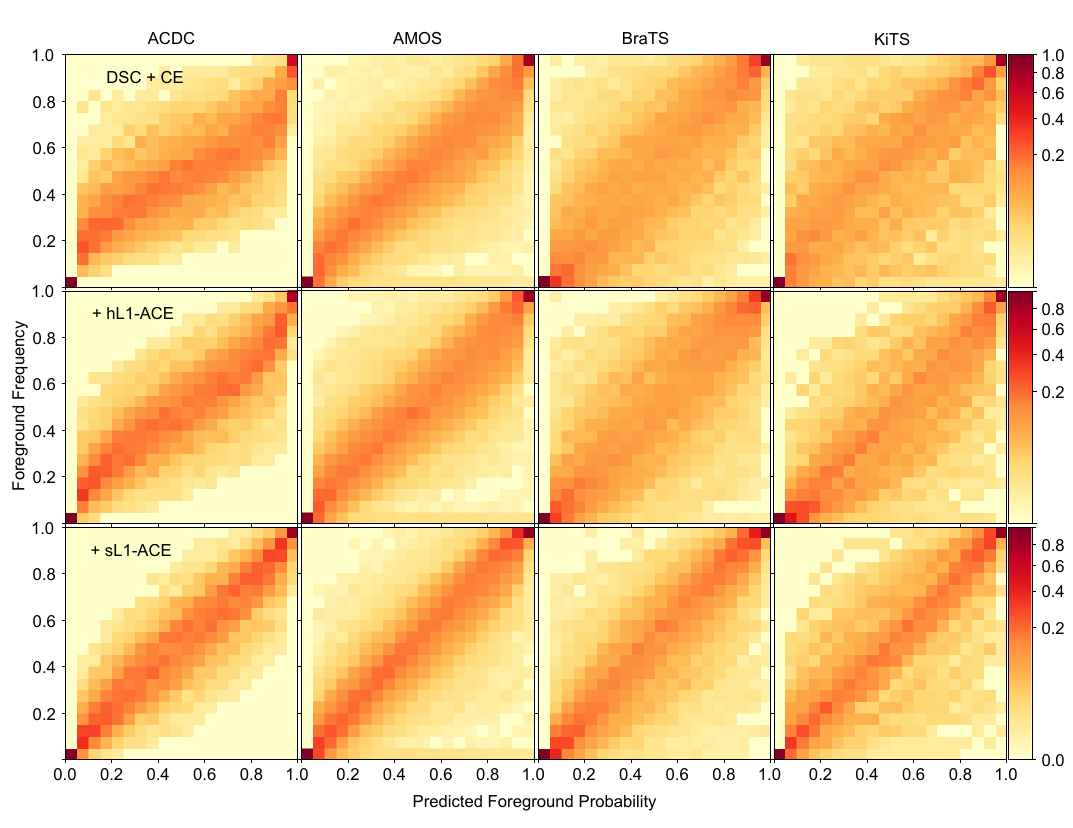}
    \caption{Dataset reliability histograms averaged over classes from each dataset, for baseline loss (DSC + CE) and baseline plus mL1-ACE with hard (hL1-ACE) and soft (sL1-ACE) binning. Gamma correction is applied to the histograms for better visualisation of lower frequencies.}
    \label{fig:dataset_reliability_diagram}
\end{figure*}

The observed improvements in calibration resulting from the inclusion of auxiliary mL1-ACE losses are further supported by the dataset reliability histograms in  \figref{fig:dataset_reliability_diagram}.
These histograms show a narrowing of the distribution of predicted probabilities towards the true class frequency, indicating improved calibration, across all datasets, with the greatest improvement resulting from the use of soft-binned mL1-ACE.

\subsection{Segmentation Probability Maps}
\begin{figure*}[htbp]
    \centering
    \includegraphics[width=\textwidth]{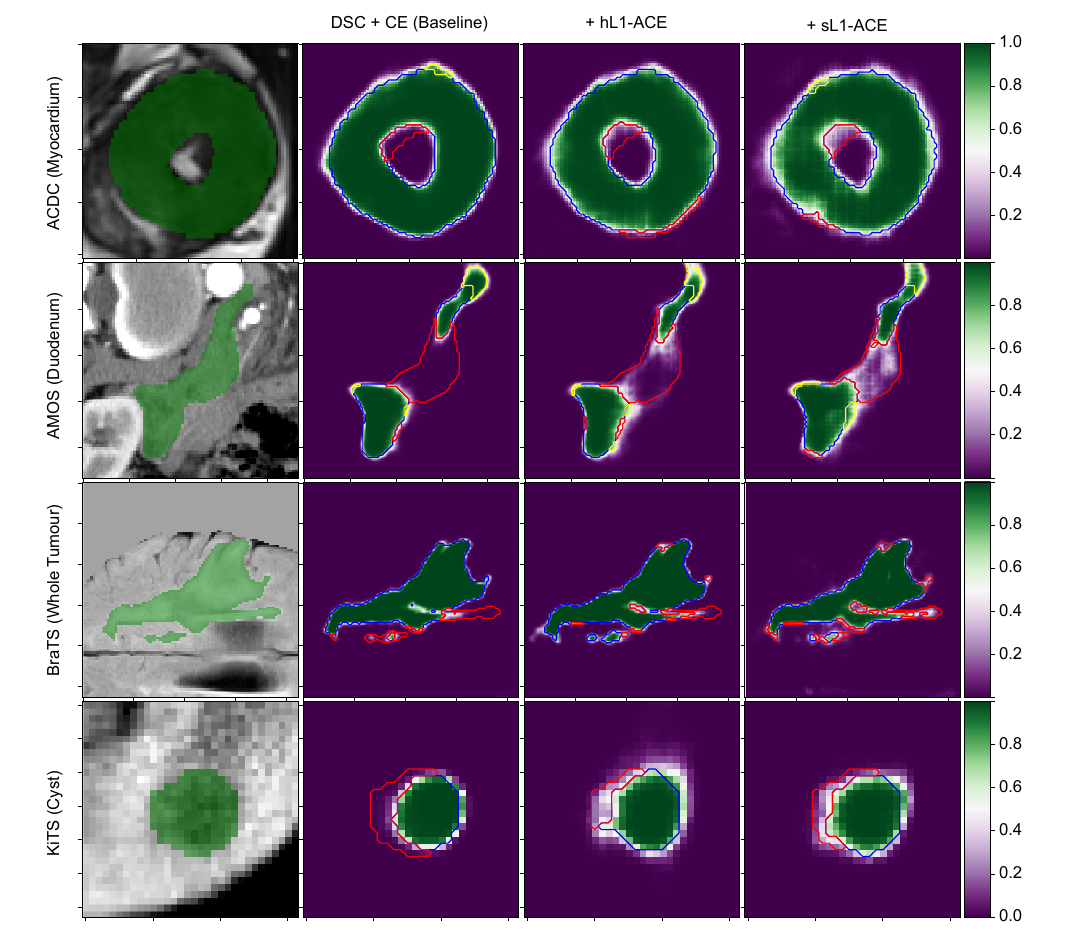}
    \caption{Comparison of predicted segmentation probability maps between baseline and models trained with our auxiliary calibration losses. The leftmost column shows input images with ground truth segmentations. Rows represent different anatomical classes from selected dataset cases. Overlaid contours highlight prediction alignment with ground truth: true positives in blue, false negatives in red, and false positives in yellow.
    }
    \label{fig:segmentation_results}
\end{figure*}

We visualise the resultant probability maps from the models trained with the baseline and mL1-ACE losses in  \figref{fig:segmentation_results}, for a given case and class from the testing dataset.
These maps show the predicted foreground probability for each voxel for a given slice, with contours overlaid to show the relationship between the discrete prediction and the ground truth.
We observe that the mL1-ACE losses result in a more diffuse probability distribution, with a softening of the class probabilities towards the decision boundary for the false positive (yellow) and particularly the false negative (red) regions of the segmentation.
This indicates that the mL1-ACE losses are able to improve the calibration of the models, reducing the overconfidence in the model predictions.
All segmentation probability maps can be found on our public Github repository, including those from the 10 worst DSC cases for each dataset.
\subsection{Sensitivity Studies}

\begin{figure*}[htbp]
    \centering
    \includegraphics[width=0.95\textwidth]{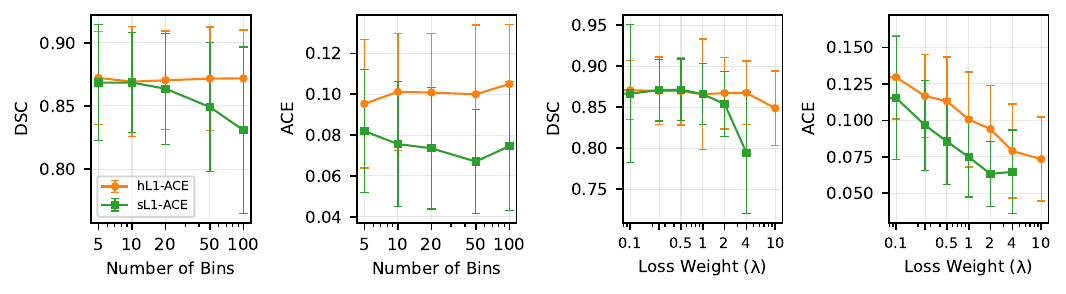}
    \caption{Sensitivity study on the ACDC dataset for hard-binned (hL1-ACE) and soft-binned (sL1-ACE) mL1-ACE variants, showing segmentation performance (DSC) and calibration (Macro-ACE) as a function of the number of bins and the loss weighting~$\lambda$.}
    \label{fig:sensitivity_study}
\end{figure*}

\textbf{Bin sensitivity.} Figure~\ref{fig:sensitivity_study} shows DSC and Macro-ACE when varying the number of bins from 5 to 100. Hard L1-ACE exhibits near-constant DSC and only minor variation in Macro-ACE across all bin counts. In contrast, Soft L1-ACE shows a reduction in DSC as the number of bins increases, while Macro-ACE remains relatively stable for both variants.

\textbf{Loss weighting sensitivity.} Figure~\ref{fig:sensitivity_study} also reports the effect of varying the loss weighting~$\lambda$ from 0.1 to 10.0. Increasing~$\lambda$ reduces Macro-ACE for both Hard and Soft L1-ACE, indicating improved calibration.
DSC for Hard L1-ACE remains stable over most of this range, with noticeable degradation only at the highest weighting. For Soft L1-ACE, DSC decreases more markedly as~$\lambda$ increases, particularly for $\lambda \geq 1.0$.

\section{Discussion and Conclusions}
\label{sec:discussion}

\subsection{Limitations of Expected Calibration Error}
While Expected Calibration Error (ECE) has been widely adopted as a metric for assessing calibration performance, our work highlights several limitations of relying on ECE as a singular measure.
In particular, ECE is heavily influenced by the distribution of counts in each bin, with the first and last bins often being overrepresented when using modern DNNs, as shown in \figref{fig:reliability_diagram_example}.
As a result, ECE tends to be insensitive to the few but clinically relevant errors that modern deep neural networks make, a limitation that becomes especially pronounced in semantic segmentation tasks, where the output space consists of a large number of probability estimates.
This issue is further exacerbated in 3D segmentation, such as in medical imaging applications where images can contain over $10^8$ voxels, with most of  it belonging to the background class.
To address these challenges, we propose the use of Average Calibration Error (ACE) as a more robust alternative.
Unlike ECE, ACE weights all bins equally, thereby offering more pronounced calibration near the decision boundary (0.5).
Although Maximum Calibration Error (MCE) provides insight into the worst-case scenario across bins, our observations indicate that MCE and ACE are highly correlated, supporting the use of ACE as a reliable calibration metric.

\subsection{Micro- versus Macro-averaging}
Our analysis reveals that micro-averaged calibration errors are consistently lower than their macro-averaged counterparts.
This outcome is expected as micro-averaging aggregates binning data across all cases, allowing for miscalibration errors from different inputs to cancel each other out for a given bin.
Although it remains an open question as to whether macro- or micro-averaging is more appropriate for semantic segmentation tasks, we report both metrics to align with established practices in the classification literature where micro-averaging is used.
Importantly, we find that the improvements in calibration achieved through our auxiliary calibration losses are more pronounced when assessed with micro-averaged metrics, despite their formulation essentially being macro-averaging.

\subsection{Presence of Missing Ground Truth Classes}
Another critical observation is the significant impact that images with missing ground truth classes can have on calibration performance.
During both training and evaluation, the inclusion of cases with missing classes tends to skew image-level calibration error metrics, leading to an apparent decrease in calibration improvement.
By excluding missing classes from the calibration error computation during training, as we have presented here, a more consistent improvement in calibration performance is observed.
Similarly, by limiting evaluation to cases with full ground truth classes we observe a more consistent improvement in calibration metrics, however, for completeness we primarily report results on the entire testing dataset.
The softening of probability distributions towards the decision boundary, an effect induced by our auxiliary losses, can inadvertently penalise predictions when a class is entirely absent.
This observation represents a limitation of our method and suggests that the utility of these auxiliary losses may be reduced in scenarios where missing classes are prevalent.
Although this issue is not specific for our calibration losses, as it is also shown to affect DSC loss~\cite{Tilborghs2022}.


\subsection{Comparison of Hard and Soft Binning}
When comparing the hard- and soft-binned calibration losses, our results indicate that the soft-binned variant generally delivers superior calibration performance, but at the cost of a statistically significant reduction in segmentation performance (DSC) on all datasets except KiTS.

Quantitatively, hard-binning (hL1-ACE) provides moderate calibration improvements (average 16\% ACE reduction) while essentially maintaining segmentation performance (average +0.03 percentage points DSC). In contrast, soft-binning (sL1-ACE) achieves larger calibration gains (average 33\% ACE reduction) but with a small decrease in segmentation accuracy (average -0.7 percentage points DSC), highlighting a systematic calibration–accuracy trade-off.

Conversely, the hard-binned calibration loss maintains segmentation performance comparable to the baseline, with only a modest DSC decrease on AMOS 22, suggesting that it may be more appropriate in applications where segmentation accuracy is paramount.
The choice between variants should depend on application requirements: hard-binning is preferable when segmentation accuracy is paramount and some calibration improvement is desired; soft-binning is preferable when calibration is critical (e.g., for safety-critical applications, active learning, or quality control) and the small segmentation trade-off is acceptable.

Our findings align with recent analyses of the calibration-generalisation trade-off~\cite{Chidambaram2025Reassessing}, which demonstrate that improved calibration and segmentation accuracy are inherently in tension. Rather than simply ``improving reliability'', our auxiliary losses provide practitioners with explicit control over this trade-off: Hard mL1-ACE offers a conservative option that maintains segmentation performance with moderate calibration gains, while Soft mL1-ACE enables stronger calibration improvements when the application justifies accepting a small reduction in DSC. This framing acknowledges that calibration improvements are coupled to, and sometimes limited by, generalisation behaviour by design.

The sensitivity analysis further clarifies these trade-offs. Hard L1-ACE is comparatively robust to both bin count and loss weighting: DSC remains stable across a wide range of settings, while calibration steadily improves as the ACE weighting increases, making it a suitable choice when segmentation accuracy is the primary objective and hyperparameter tuning capacity is limited. Soft L1-ACE, in contrast, achieves stronger calibration improvements but is more sensitive to hyperparameters: larger numbers of bins and higher ACE weightings lead to noticeable reductions in DSC, reflecting a sharper accuracy–calibration trade-off and suggesting the use of coarser binning and moderate or lower ACE weightings (e.g., $\lambda < 1.0$) when segmentation quality must be preserved.

The dynamics of joint optimisation explain why calibration and segmentation performance are inherently in tension. Overlap metrics such as DSC reward sharp, confident predictions at correct spatial locations, whereas calibration losses are spatially agnostic and depend only on the alignment between predicted confidence and empirical accuracy. The soft-binned variant achieves superior calibration because its overlapping bin assignments (\figref{fig:binning_comparison}) impose stronger smoothing on class probabilities, more effectively reducing overconfidence but also blurring decision boundaries and thereby degrading DSC.
The hard-binned variant, with its discrete binning, preserves sharper probability distributions and thus segmentation accuracy, while still improving calibration through explicit optimisation of the ACE objective.
In practice, this means that Hard L1-ACE is a robust default for general use, whereas Soft L1-ACE is best deployed when improved calibration justifies more careful hyperparameter tuning and a modest reduction in DSC.

We hypothesise that this behaviour is most likely explained by an \emph{accumulation of contributions} in the soft-binning formulation, rather than by discontinuities, which soft kernels mitigate. Concretely, with hard binning, each voxel contributes to exactly one bin, whereas with soft binning each voxel contributes to multiple neighbouring bins. As the bin width shrinks (high bin count), the number of overlapping bin assignments per voxel does not decrease, but the per-bin gradients can become sharper, so the \emph{aggregate} calibration-loss gradient can increase in magnitude and variability. This may over-regularise predicted probabilities, reducing sharpness/confidence, and thereby degrade DSC even when summary calibration metrics remain stable. Practitioners should therefore prefer fewer bins (e.g., 10--20) when using soft-binned mL1-ACE to avoid this effect.

\subsection{Comparison with Temperature Scaling and Cross-Entropy Only}
Post-hoc temperature scaling provides a simple and effective way to improve calibration, and our results confirm it achieves comparable calibration to Hard mL1-ACE while preserving DSC identically.
Train-time calibration offers several complementary advantages: (1) This allows for full-dataset training without holding back samples for calibration, a critical advantage in data-scarce medical imaging contexts; (2) it enables the model to learn calibration-aware representations during training; (3) it can be combined with post-hoc methods for additional gains; (4) it provides flexibility to adjust the calibration-accuracy trade-off through hyperparameter selection; and (5) it allows calibration to vary across spatial locations and anatomical contexts, which may be beneficial in dense prediction tasks where confidence miscalibration is not uniform across an image. However, when simplicity is paramount and a calibration set is available, temperature scaling remains a strong choice.
Local Temperature Scaling~\cite{ding2020local} may offer further post-hoc improvements for dense prediction tasks, but such methods still require a separate calibration phase, whereas our mL1-ACE losses integrate calibration directly into the training objective.

Cross-entropy alone produces better-calibrated probabilities than DSC+CE but does so at the expense of segmentation accuracy, illustrating the well-known trade-off between these objectives \cite{lossodyssey2021, Mehrtash2020confidence}.
Both hard and soft mL1-ACE variants achieve superior calibration compared to CE-only while preserving the strong segmentation performance provided by DSC.
This demonstrates that our auxiliary losses provide explicit control over the calibration–accuracy trade-off, enabling practitioners to select the variant and hyperparameters that best suit their application requirements.

\subsection{Practical Implications}
Our work demonstrates that the use of our auxiliary losses makes the class-wise probabilities more reliable as indicators of accuracy.
While we cannot definitively quantify the extent to which a given reduction in calibration error translates to improved clinical outcomes without further downstream studies, the improved reliability of confidence estimates has several potential practical applications.
Models trained with auxiliary calibration losses could be integrated into quality control systems to automatically flag uncertain predictions that are likely to be inaccurate, thereby streamlining the review process for radiologists.
Furthermore, reliable uncertainty estimates are crucial for active learning scenarios, where they can guide sample selection to maximise data efficiency.
Finally, providing clinicians with trustworthy confidence information alongside segmentations can enhance trust in AI systems and support more informed decision-making.
However, further studies are needed to verify these impacts in clinical settings.

\subsection{Limitations of Calibration Losses}
Although our auxiliary calibration losses lead to marked improvements in calibration metrics, it remains to be demonstrated that these gains translate to better clinical outcomes.
It is perhaps not surprising that losses based on established calibration metrics would improve those metrics during evaluation.
However, our approach is notable for allowing these metrics to be computed at an image level and directly incorporating these as auxiliary losses in a semantic segmentation context.
This strategy leverages the large capacity of deep neural networks to learn better-calibrated predictions while exerting only limited negative effects on segmentation performance.
One inherent limitation of our current method is that the process of binning foreground probabilities strips away spatial information, thereby limiting the model's ability to capture spatial correlations in uncertainty and confidence.
Although the visualisations in \figref{fig:segmentation_results} demonstrate instances where improvements in calibration align with error regions, this correspondence is not consistent across all cases.

A further avenue for future work concerns the binning kernels themselves. In this study, we employed fixed square and triangular kernels for hard- and soft-binning, respectively. Learning the kernel shape during training, for example, by parameterising the bin membership functions and optimising them jointly with the network weights, similar to~\cite{Karandikar2021SoftCalib}, could allow the model to adapt to dataset-specific probability distributions.

Recent work has highlighted important considerations for calibration in deep learning. Chidambaram \& Ge discuss the calibration-generalisation trade-off and methods for fair comparison of calibration approaches~\cite{Chidambaram2025Reassessing}. Wang \etal argue that some degree of overconfidence may be acceptable or even beneficial in certain contexts~\cite{Wang2021Rethinking}. Our work contributes to this discussion by providing practitioners with explicit control over the calibration objective, enabling them to navigate the accuracy-calibration trade-off based on their application's specific requirements.

Our evaluation across four diverse datasets, covering different modalities (MRI, CT) and anatomical regions, provides strong evidence of our method's reliability. Dedicated cross-dataset generalisation experiments, where models trained on one dataset are evaluated on another, would be a valuable direction for future work to further assess robustness under severe distribution shifts.

\subsection{Conclusion}
In this study, we introduced differentiable formulations of marginal L1 Average Calibration Error (mL1-ACE) as auxiliary losses specifically designed for improving the calibration of deep neural networks in medical image segmentation tasks.
Through comprehensive evaluations on four publicly available datasets (ACDC, AMOS, BraTS, KiTS), we demonstrated that integrating calibration-aware losses at training time significantly reduces calibration errors, as evidenced by substantial improvements in both macro- and micro-averaged ACE and MCE metrics.
The soft-binned mL1-ACE variant provided the largest calibration improvements, reducing macro- and micro-averaged ACE by an average of 33\% and 51\%, respectively, at a small cost in segmentation accuracy (-0.7\% DSC). In contrast, the hard-binned variant maintained segmentation accuracy (+0.03\% DSC) while reducing macro- and micro-averaged ACE by 16\% and 32\%, respectively.

Importantly, our comprehensive baseline comparisons demonstrate that both mL1-ACE variants outperform not only the standard DSC+CE baseline but also CE-only training in terms of calibration, across almost all metrics. While CE alone produces better-calibrated predictions than DSC+CE, this comes at a substantial cost to segmentation performance.
Our approach provides practitioners with explicit control over this fundamental trade-off: hard-binned mL1-ACE offers a conservative option that maintains segmentation accuracy while improving calibration, whereas soft-binned mL1-ACE enables stronger calibration gains when the application justifies a modest reduction in DSC. This flexibility represents a significant practical advance for medical imaging applications, where different deployment scenarios may prioritise accuracy or calibration differently.

Our analysis also highlights limitations in standard calibration metrics, such as Expected Calibration Error (ECE), advocating instead for the use of Average Calibration Error (ACE) to better capture miscalibration, particularly in boundary regions critical to clinical decision-making.
Additionally, we introduced dataset reliability histograms, providing intuitive visual insights into calibration performance across entire datasets.

Despite these advances, we identified challenges posed by images with missing ground-truth classes in datasets, suggesting future directions towards more robust handling of such scenarios.
Overall, our findings underscore the importance and feasibility of directly optimising for calibration during training, contributing significantly to the development of more reliable and trustworthy medical imaging AI systems suitable for clinical integration.

\section*{Acknowledgments}
This work was supported by EPSRC CDT [EP/S022104/1], Intel, Medtronic/RAEng [RCSRF1819\textbackslash7\textbackslash34], and Wellcome/EPSRC [WT203148/Z/16/Z; NS/A000049/1]

\bibliographystyle{plain}
\bibliography{references}

\end{document}